\documentclass{article} 
\usepackage{iclr2020_conference,times}

\usepackage{amsmath,amsfonts,bm}

\def\eqref#1{equation~\ref{#1}}

\def\1{\bm{1}}

\DeclareMathAlphabet{\mathsfit}{\encodingdefault}{\sfdefault}{m}{sl}
\SetMathAlphabet{\mathsfit}{bold}{\encodingdefault}{\sfdefault}{bx}{n}

\DeclareMathOperator{\sign}{sign}

\usepackage{hyperref}
\usepackage{url}

\usepackage{graphicx}
\usepackage{subfigure}
\usepackage{amssymb}
\usepackage{amsmath}
\usepackage{amsthm}
\usepackage{gensymb}

\newtheorem{theorem}{Theorem}
\newtheorem{lemma}{Lemma}

\newcommand{\xy}{\begin{bmatrix}x_0\\x_1\end{bmatrix}}

\title{A Simple Geometric Proof for the Benefit of Depth in ReLU Networks}

\author{Asaf Amrami$^{1,2}$ \& Yoav Goldberg$^{1,2}$ 
\\
$^1$ Bar Ilan University\\
$^2$ Allen Institute for Artificial Intelligence\\
}

\iclrfinalcopy 
\begin{document}

\maketitle

\begin{abstract}
We present a simple proof for the benefit of depth in multi-layer feedforward network with rectified activation (``depth separation''). Specifically we present a sequence of classification problems indexed by $m$ such that (a)  for any fixed depth rectified network there exist an $m$ above which classifying problem $m$ correctly requires exponential number of parameters (in $m$); and (b) for any problem in the sequence, we present a concrete neural network with linear depth (in $m$) and small constant width ($\leq 4$) that classifies the problem with zero error.

The constructive proof is based on geometric arguments and a space folding construction.

While stronger bounds and results exist, our proof  uses substantially simpler tools and techniques, and should be accessible to undergraduate students in computer science and people with similar backgrounds.

\end{abstract}

\section{Introduction}

We present a simple, geometric proof of the benefit of depth in deep neural networks. 

We prove that there exist a set of functions indexed by $m$, each of which can be efficiently represented by a depth $m$ rectified MLP network requiring $O(m)$ parameters. However, for any bounded depth rectified MLP network, there is a function $f_m$ in this set that cannot be represented by the network unless it has exponential number of parameters in $m$. 

More formally,
We will prove the following theorem:
\begin{theorem}[Depth Separation]
\label{theorem:main}
There exists a sequence of functions $f_1,f_2,... : \mathbb{R}^2 \mapsto \{-1,1\}$
such that:
\begin{enumerate}
    \item[a] [Bounded depth network is exponential in size] For any rectified MLP with bounded depth $d$, 
    solving problem $f_m$ requires a width of at least $b^m$ to solve, where $b>1$ is a constant determined by the bounded depth $d$.
    \item[b][Utility of depth]
    For any problem $f_m$ there is a rectified MLP with linear number of parameters in $m$ that solves $f_m$ perfectly. More concretely, there exists a network with depth $m+2$ and layer width $\leq 4$ that perfectly represents $f_m$. 
\end{enumerate}
\end{theorem}

While this is not a novel result,
a main characteristic of our proof is its simplicity. In contrast to previous work, our proof uses only basic algebra, geometry and simple combinatorial arguments. As such, it can be easily read and understood by newcomers and practitioners, or taught in a self-contained lecture in an undergraduate class, without requiring extensive background.
Tailoring to these crowds, our presentation style is more verbose then is usual in papers of this kind, attempting to spell out all steps explicitly. We also opted to trade generality for proof simplicity, remaining in input space $\mathbb{R}^2$ rather than the more general $\mathbb{R}^n$, thus allowing us to work with lines rather than hyperplanes. Beyond being easy to visualize, it also results in somewhat simpler proofs of the different lemmas.

\section{Related Work}
The expressive power gained by depth in multi-layer perceptron (MLP) networks is relatively well studied, with multiple works showing that deep MLPs can represent functions that cannot be represented by similar but shallower networks, unless those have a significantly larger number of units \citep{delalleau2011shallow,pascanu2013number,bianchini2014complexity}.

\citet{telgarsky2015representation,telgarsky2016benefits} show that network depth facilitate fast oscillations in the network response function. Oscillations enabled by a linear growth in depth are shown to require exponential growth in the number of units when approximated well by a shallower network.

\citet{eldan2016power} study approximation to the unit sphere in a wide family of activation function. In their construction they show that a 3-layer MLP could first compute the polynomial $x^2$ for each of the dimensions and use the last layer to threshold the sum of them to model the unit sphere indicator. They analytically show that the same approximation with 2-layer network requires exponentially growth in width with precision.

\citet{yarotsky2017error,DBLP:journals/corr/SafranS16}, show that depth is useful for approximating polynomials by ReLU MLPs. Specifically, that $f(x)=x^2$ could be efficiently approximated with network depth.

While results similar to one presented here could be derived by a combination of the construction in \citet{eldan2016power} and the polynomial approximation of \citet{yarotsky2017error}, we present a different (and to our taste, simpler) proof, using a geometric argument and a bound on the number of response regions of ReLU networks, without explicitly modeling the $x^2$ polynomial.

The ReLU MLP decision space was studied by \cite{pascanu2013number}. They show that the input space is sequentially refined by the ReLU and linear operations of the network to form separated convex polytopes in the input space. They call these regions \emph{response regions}. They also establish a lower bound on the maximal number of regions, a bound which is tightened by \citet{montufar2014number,raghu2017expressive,arora2016understanding,serra2017bounding}. 
We rely on the notion of response region in our proof, while attempting to provide an accessible explanation of it. Some of the lemmas we present are simplified versions of results presented in these previous works.

\section{Background}
\subsection{Linearity and piecewise linearity. Convexity.}

A \emph{linear function} is a function of the form $f(x) = \mathbf{A}x + \mathbf{b}$. For affine spaces (like the Euclidean space), this is also called an \emph{affine transformation} of the input. In a \emph{piecewise linear} function the input space is split into regions, and each region is associated with a linear function.
A composition of linear functions is linear. A composition of piecewise linear functions is piecewise linear.

A $2d$ region is \emph{convex} iff, for any two points in the region, all points on the line connecting the two points is also within the region. A polygon with all internal angles $<180^o$ is a convex region.

\subsection{ReLU MLP with d Layers}
A ReLU MLP with $d$ layers parameterized by $\Theta$ is a multivariate function defined as the composition: $$F(X;\Theta)= h^{out} \circ h_d^A \circ \sigma \circ h_{d-1}^A \circ \sigma \ldots \circ h_{2}^A \circ \sigma \circ h_{1}^A(X)  $$
Where $h^A_i$s are parameterized affine transformations; $\Theta$ the set of parameters in them; and 
$\sigma$ is the ReLU activation function: a non linear element-wise activation function defined by $\sigma(x)=max\{0,x\}$.
We consider ReLU MLPs where all hidden layers have the same width $w$.\footnote{This subsumes networks with layers with width $<w$, as these are equivalent to width $w$ layers with zeroes in specific regions of the parameters.} Without loss of generality we define the last layer of network, $h^{out}$, as a weighted sum over its inputs where a sum strictly greater than zero is mapped to the $1$ class, and otherwise to the $-1$ class.\footnote{In common ML operations, this simply means that $h^{out}$ multiplies by a vector and takes the sign of the resulting scalar.}

The combination of linear operations and the ReLU function result in a piecewise linear function of the input $X$.

\subsection{ReLU MLP response regions}
Piecewise linear activation functions such as ReLU split the input space into convex regions of linear activation. This is asserted formally and visualized in \cite{hanin2019complexity}.
The ReLU function has two regions (``pieces'') of linearity $x>0, x\leq0$. Within each of these, linearity is maintained. The sequential composition of affine transformations and the ReLU operations created by the MLP layers, divides the the input space into convex polytopes (in $2D$, as we consider here, these are convex polygons). 
Within each such polytope, the function behaves linearly. We call these polytopes \emph{linear response regions}.

The number of these linear response regions, and specifically the effect of MLP depth on the maximal number of regions, was studied in multiple works \cite{montufar2014number,raghu2017expressive,arora2016understanding,serra2017bounding}. We focus on the simpler case of 2-class classification ReLU MLP on the Euclidean plane and denote the maximal number of response regions of a network of $d$ layers each with $w$ units as $r(w,d)$. 

Our presentation of the proof of lemma \ref{lemma-max-regions-bounded-depth} gives more insight into response regions.

\subsection{Folding transformations}
\label{folding}
\citet{montufar2014number} present the concept of folding transformation and their implementation with ReLUs. Looking at one or more layers as a function $f:\mathbb{R}^2 \to \mathbb{R}^2$, a folding transformation maps a part of the input space to coincide with another. 
Subsequent operations on the resulting space will apply to both parts, indifferently to their origin in their initial position. 
As a simple example, consider a ReLU MLP of input dimension 1. A simple folding two-layer transformation could easily model the function $abs(x)=|x|$, mapping the negative input values to their positive counterparts.\footnote{This is achieved by linearly mapping $x$ into the pair $[x,-x]$, and then applying ReLU and summing ($abs(x) = ReLU(x) + ReLU(-x) = $).} Afterwards, any operation in subsequent layers will apply to both the negative values and positive values. This simple mechanism of ``code reuse" is key to our constructed deep network and its unit-efficiency.
Intuitively, our construction resembles children paper-cutting, where a sheet of paper is folded multiple times, then cut with scissors. Unfolding the paper reveals a complex pattern with distinctive symmetries.
Tracing and cutting the same pattern without any paper folding would require much more effort.
Analogously, we'll show how deep networks could implement ``folds'' through their layers and how ReLU operations, like scissor cuts, are mirrored through the symmetries induced by these folds. Conversely, shallow networks, unable to ``fold the paper'', must make many more cuts --- i.e. must have much more units in order to create the very same pattern.\footnote{\citet{malach-shalev-fractals} show a similar construction using Fractals.}

\section{Main Proof}

\begin{figure*}[h!]
    \centering
    \includegraphics[width=13cm]{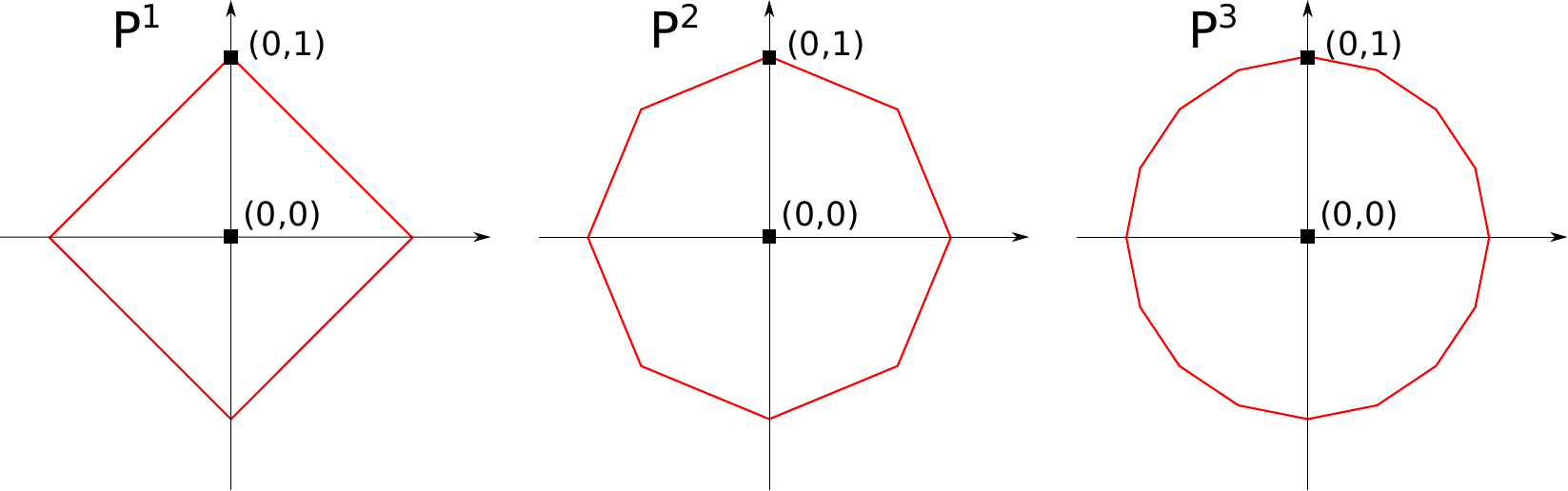}
    \caption{The problem family $f_m$ is characterized by regular polygons, where polygon $P_m$ has $2^{m+1}$ edges.}
    \label{fig:prob}
\end{figure*}

\subsection{The problems $f_m$}

Let $P_m$ be a regular polygon with $2^{m+1}$ edges (Figure \ref{fig:prob}).\footnote{A regular polygon is a polygon that is both equi-angular (whose angles are all equal) and equilateral (whose edges are all equal).} Without loss of generality, $P_m$ is centered around the origin, bounded by the unit circle, and has a vertex at $(0,1)$.\footnote{Any other regular polygon can be shifted, rotated and scaled to these conditions using affine transformation with $O(1)$ parameters.}
The set of polygons $P_1,P_2,...$ approaches the unit circle as $m \rightarrow \infty$.
Let $f_m$ be the function with decision boundary $P_m$:
\[
f_m(x) = \begin{cases}
1 & x \text{ is inside } P_m\\
-1 & \text{otherwise}
\end{cases}
\]

Points within polygon $P_m$ are of class $1$, while other points are of class $-1$. 

\subsection{A bounded-depth network representing $f_m$ must be exponentially wide.}

We begin with proving (a) of Theorem 1. We will use the following lemmas, with proofs for the lemmas provided in the next section.

\begin{lemma} A rectified MLP is a piecewise linear function.
\label{lemma-picewislin}
\end{lemma}
Proof: A linear layer followed by a rectifier is piecewise-linear. A composition of piecewise linear functions is itself piecewise linear.

\begin{lemma}
\label{lemma-response-regions-problem}
Modeling $f_m$ as a piecewise linear function requires at least $2^m$ response regions.

\end{lemma}

\begin{lemma}
\label{lemma-max-regions-bounded-depth}
Rectified MLP with input in $\mathbb{R}^2$, with $d$ hidden layers and maximal layer width of $w$, has at most $w^{2d}$ response regions.
\end{lemma}

Together, Lemma \ref{lemma-response-regions-problem} and Lemma \ref{lemma-max-regions-bounded-depth} show how network width $w$ behaves when the problem grows more complex. 
To prove Theorem (\ref{theorem:main}a), we need to show that $w$ is exponential. Namely, we will show that there is a base $b > 1$ such that $w\geq b^m$.
From Lemma \ref{lemma-response-regions-problem}, modeling $f_m$ requires $2^m$ response regions. From lemma \ref{lemma-max-regions-bounded-depth}, a network with depth $d$ has at most $w^{2d}$ regions. To model $f_m$, we thus need $w^{2d} \geq 2^m$ response regions.

Taking the $2d$ root from both sides\footnote{Both sides' variables are strictly positive, allowing taking roots and logarithms.} we get $w \geq 2^{\frac{m}{2d}} = (2^{\frac{1}{2d}})^m$. Since the depth $d$ is constant, denote $b=2^\frac{1}{2d} > 1$, leading to $w\geq b^m$ as desired. This concludes the proof of Theorem (\ref{theorem:main}a). An alternative view of the same math which may be simpler to some readers: we can re-write $w$ as $2^{\log_2w}$, leading to $(2^{\log_2w})^{2d} = 2^{\log_2w\cdot2d} \geq 2^m$. Obtaining $log_2w\cdot2d \geq m$, where the logarithm on the left indicates that we require an exponential growth in $w$ to match $m$ for a fixed depth $d$ as $m$ grows.

\subsection{Efficient depth-$m$ solution exists.}
Lemma \ref{lemma-max-regions-bounded-depth} provides a lower bound for the size of any zero error network. We now turn to prove Theorem (\ref{theorem:main}b) by showing how to construct a linear depth and bounded width network. Our construction is based on folding transformations.

As discussed in \S\ref{folding}, we construct the regular polygon decision boundary for polygon $P_m$ through exploitation of symmetry and folding transformations.

Formally, our deep network operates as follows:  first, it folds across both the $X$ and $Y$ axes, mapping the input space into the first quadrant $(x,y) \mapsto (|x|,|y|)$. It now has to deal only with the positive part of the decision boundary. It then proceeds in steps, in which it first rotates the space around the origin until the remaining decision boundary is symmetric around the $X$ axis, and then folds around the $X$ axis, resulting in half the previous decision boundary, in the first quadrant. This process continues until the decision boundary is a single line, which can be trivially separated. The first step cuts the number of edges in the decision boundary by a factor of four, while each subsequent rotate + fold sequence further cuts the number of polygon edges in half.

\begin{figure*}[h!]
    \centering
    \includegraphics[width=13cm]{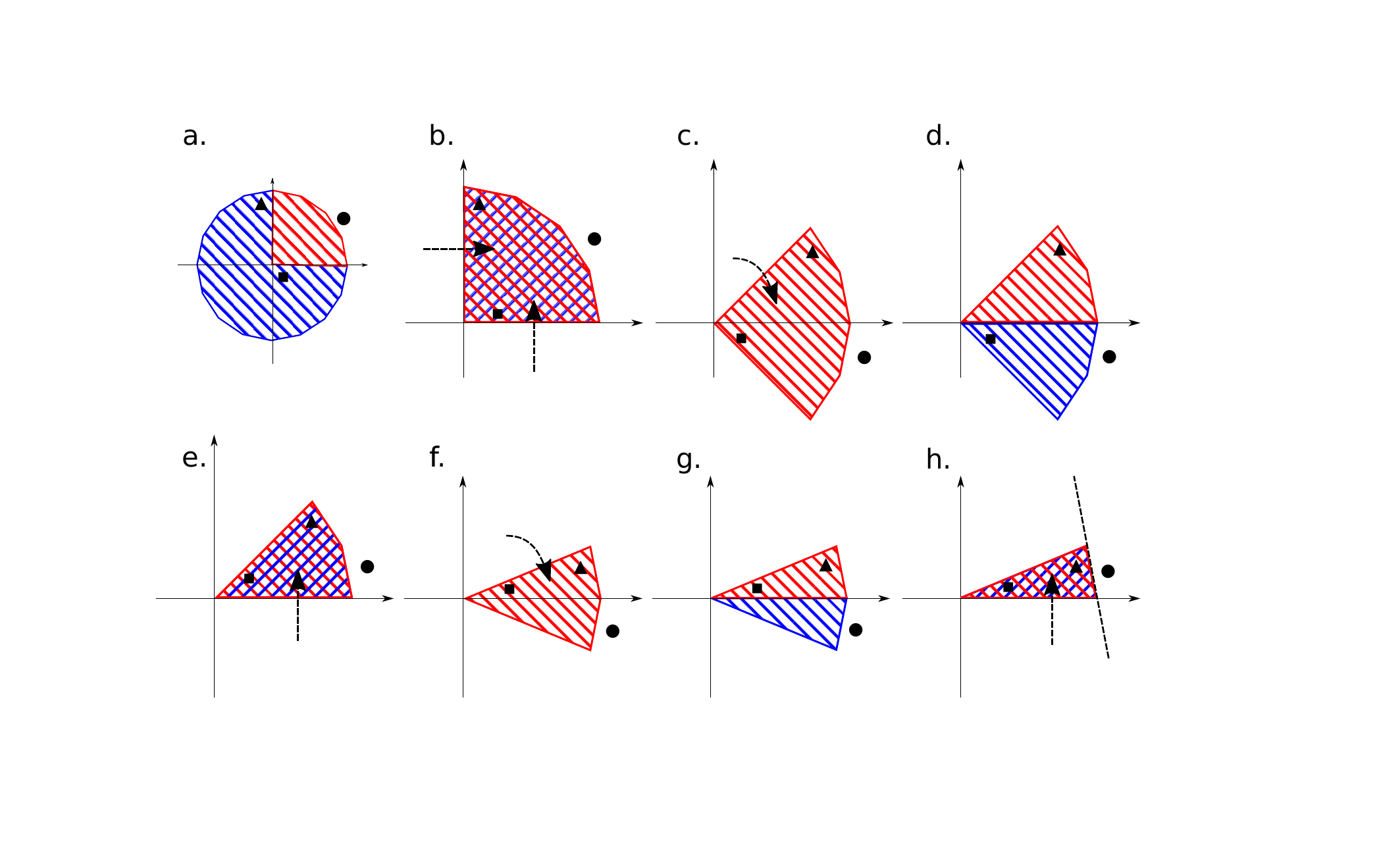}
    \caption{ Constructing $P_3$ using folding and rotation transformations.
    The 3 blackened markers show how 3 points in the input space are transformed during this process. 
    (a-b) a $\textit{foldXY}$ operation maps all points to the first quadrant. (c) the slice is rotated clockwise by 45\degree using a linear transformation. (d-e) the bottom half is mapped into the first quadrant using a $\textit{foldX}$ operation. (f) rotate by $45/2\degree$. (g-h) folding. final rotation by $45/4\degree$ and a final linear decision boundary that correctly classifies the three points.}
    \label{fig:fold}
\end{figure*}

This process is depicted in Figure \ref{fig:fold}.

More formally, we require four types of transformations:

\begin{itemize}
    \item $\textit{foldXY}(\xy):\mathbb{R}^2 \to \mathbb{R}^2$ --- initial mapping of input to the first quadrant.
    \item $\textit{rotate}_{\Theta}(\xy): \mathbb{R}^2 \to \mathbb{R}^2 $ --- clockwise rotation around the origin by an angle of $\Theta$.
    \item $\textit{foldX}(\xy):\mathbb{R}^2 \to \mathbb{R}^2$ --- folding across the $X$ axis.
    \item $\textit{top}(\xy):\mathbb{R}^2 \to \mathbb{R}^1$ --- the final activation layer.
\end{itemize}

These operations are realized in the network layers, using a combination of linear matrix operations and ReLU activations. 
The \emph{rotate} operation is simply a rotation matrix. Rotating by an angle of $\Theta$ is realized as: 

\[
\textit{rotate}_{\Theta}(\xy) =
\left[
\begin{array}{rr}
cos(\Theta) & -sin(\Theta)   \\
sin(\Theta) &cos(\Theta) \\
\end{array}
\right ] 
\xy
\]

The initial folding across both $X$ and $Y$ axes first transforms the input $(x,y)$ to $(x,-x,y,-y)$ using a linear transformation.
It then trims the negative values using a ReLU, and sums the first two and last two coordinates using another linear operation, resulting in:
\[
\textit{foldXY}(\xy) = 
\left[
\begin{array}{rrrr}
1&1 & 0 & 0\\
0&0 & 1 & 1\\
\end{array}
\right]
\sigma(
\left[
\begin{array}{rr}
-1 & 0\\
1 & 0\\
0 & 1\\
0 & -1 \\
\end{array}
\right]
\xy
)
\]

Where $\sigma$ is the elementwise ReLU activation function.
Folding across the $X$ axes is similar, but as all $x$ values are guaranteed to be positive, we do not need to consider $-x$.
\[ 
\textit{foldX}(\xy) = 
\left[
\begin{array}{rrr}
1 & 0 & 0\\
0 & 1 & 1\\
\end{array}
\right]
\sigma(
\left[
\begin{array}{rr}
1 & 0\\
0 & 1\\
0 & -1 \\
\end{array}
\right]
\xy
)
\]

Finally, the final classification layer is:
\[
\textit{top}(\xy) = 
\sign(a\cdot x_0+b\cdot x_1+c)
\]

Composing these operations, the constructed network for problem $f_m$ has the form: 
\[
\textit{f}_{\textit{MLP}}(x)=
\textit{top} 
\circ
\textit{foldX}
\circ
\textit{rotate}_{\pi/2^{m+1}}
\circ \textit{foldX}
\circ \ldots \circ
\textit{rotate}_{\pi/8}
\circ \textit{foldX}
\circ \textit{rotate}_{\pi/4}
\circ \textit{foldXY} 
\]

Note that the angle of rotation is decreased by a factor of 2 in every subsequent $\textit{rotate}$.
The $\textit{rotate}$ and $\textit{foldX}$ transformations pair, folds input space along a symmetry axis and effectively reduces the problem by half. This results in a $\textit{foldXY}$ operation followed by a sequence of $m$ $\textit{foldX} \circ \textit{rotate}$ operations, followed by $\textit{top}$.

Marking a $\textit{fold}$ operation as $\mathbf{F}\sigma\mathbf{C}$ and a rotate operation as $\mathbf{R}$, where $\mathbf{F,C,R}$ being matrices, the MLP takes the form: $\mathbf{F\sigma C R F \sigma C R F \sigma C R F \ldots }$ where a sequence $\mathbf{CRF}$ of matrix operations can be collapsed into a single matrix $\mathbf{M}$. This brings us to the familiar MLP form that alternates matrix multiplications and ReLU activations. Overall, the network has $m+1$ non-linear activations (from $m$ $\textit{foldX}$ operations and $1$ $\textit{foldXY}$ operation), resulting in $m+1$ layers.

\begin{figure*}[h!]
    \centering
    \includegraphics[width=13cm]{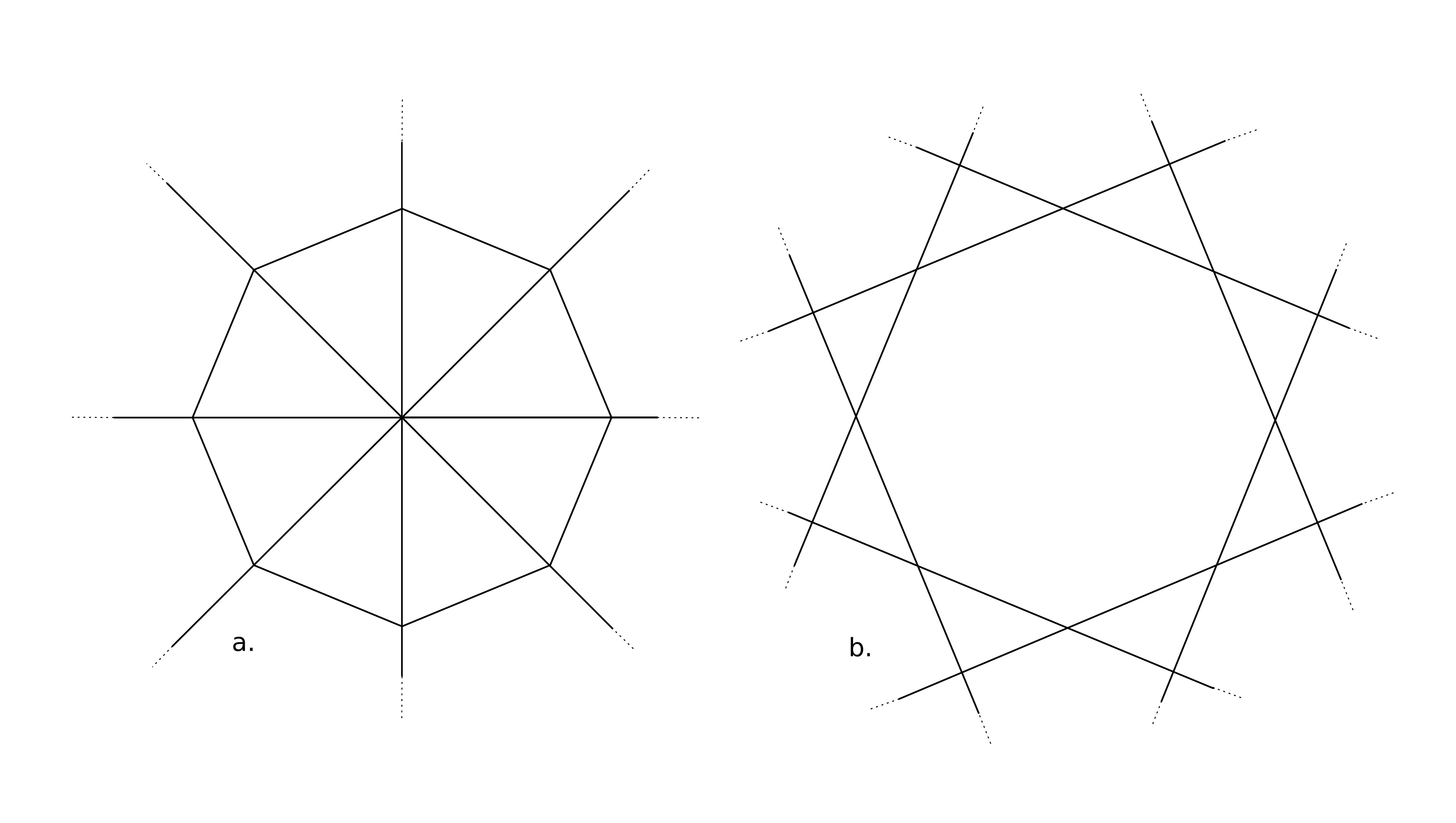}
    \caption{(a) The response regions of the constructed solution for $P_2$.  (b) A shallow, one layer MLP that solve $P_2$ - Such an MLP must model each of the regular polygon edges separately .}
    \label{fig:responses}
\end{figure*}

The response regions produced by the constructed MLP and by a shallow network are depicted in Figure \ref{fig:responses}.

\section{Proofs of Lemmas}

\subsection{Lemma \ref{lemma-response-regions-problem}}

Modeling $P_m$ as a piecewise linear function requires at least $2^m$ response regions.

\begin{figure*}[t!]
    \centering
    \includegraphics[width=13cm]{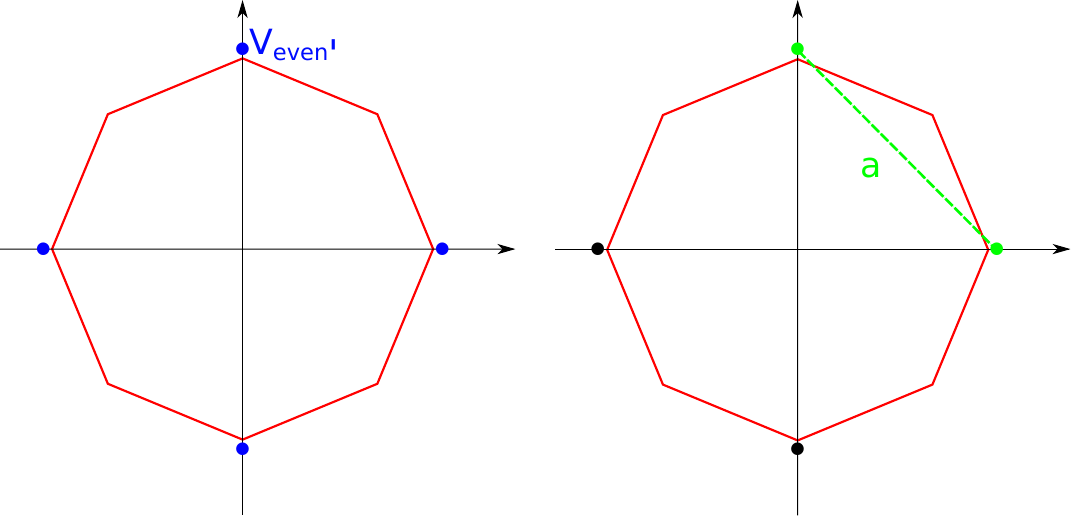}
    \caption{Left: $V_{even}'$ are created by taking every second vertex of $P_m$ then moving them slightly such that they are strictly outside $P_m$. Right: a chord $a$ in green connecting any two vertices of  $V'_{even}$, must cross $P_m$. Had both of the chord vertices been in the same response region, by convexity so do all points on $a$. By linearity, the final network activation of $a$'s points will interpolate the activation of $a$'s endpoints.}
    \label{fig:circproof}
\end{figure*}

\emph{Proof:} consider the polygon $P_m$, and let $MLP_m$ be a ReLU MLP (piecewise-linear function) correctly classifying the problem.
Let $V_{even}$ be the set of every \emph{second} vertex along a complete traversal of $P_m$.
For each vertex take an $ \epsilon $ step away from the origin to create $ V_{\textit{even}}' $ (see Figure \ref{fig:circproof}a for an illustration).
Each of the points in $V_{\textit{even}}'$ are strictly outside $P_m$ and therefore should be classified as class $-1$. 

The response regions produced by $\textit{MLP}_m$ are both convex and linear.
Let $p_i$, $p_j$ be two arbitrary points in $V_{\textit{even}}'$, $p_i \neq p_j$. We will show that $p_i$, $p_j$ belong in different response regions. Assume by contradiction that $p_i,p_j$ are in the same response region. By convexity all points in a straight line between $p_i$ and $p_j$ are also in the same response region. Also, by linearity these points have an activation value between $p_i$ and $p_j$ and therefore should also be classified as class $-1$. From the problem construction we know that lines between the even vertices of $P_m$ cross the class boundary as demonstrated in Figure \ref{fig:circproof}b. Therefore, $p_i$ and $p_j$ must lay in different response regions. Since $p_i$ and $p_j$ are arbitrary, $\textit{MLP}_m$'s number of response regions is at least $|V_{\textit{even}}'| = 2^m $.

\subsection{Lemma \ref{lemma-max-regions-bounded-depth}}
Rectified MLP with input in $\mathbb{R}^2$, with $d$ hidden layers and maximal layer width of $w$, has at most $w^{2d} = 2^{2d\log_2w}$ response regions.

\emph{Proof:} \citet{raghu2017expressive} prove a version of this lemma for input space $\mathbb{R}^n$, which have at most $O(w^{nd}) $ response regions. We show a proof for the more restricted case of inputs in $\mathbb{R}^2$, in a similar fashion. We first consider the bound for 1 hidden-layer networks, then extend to $d$ layers. The first part of the proof follows classic and basic results in computational geometry. The argument in the second part (move from 1 to $d$ layers) is essentially the same one of \citet{raghu2017expressive}.

\begin{figure}[t!]
    \centering
    \includegraphics[width=8cm]{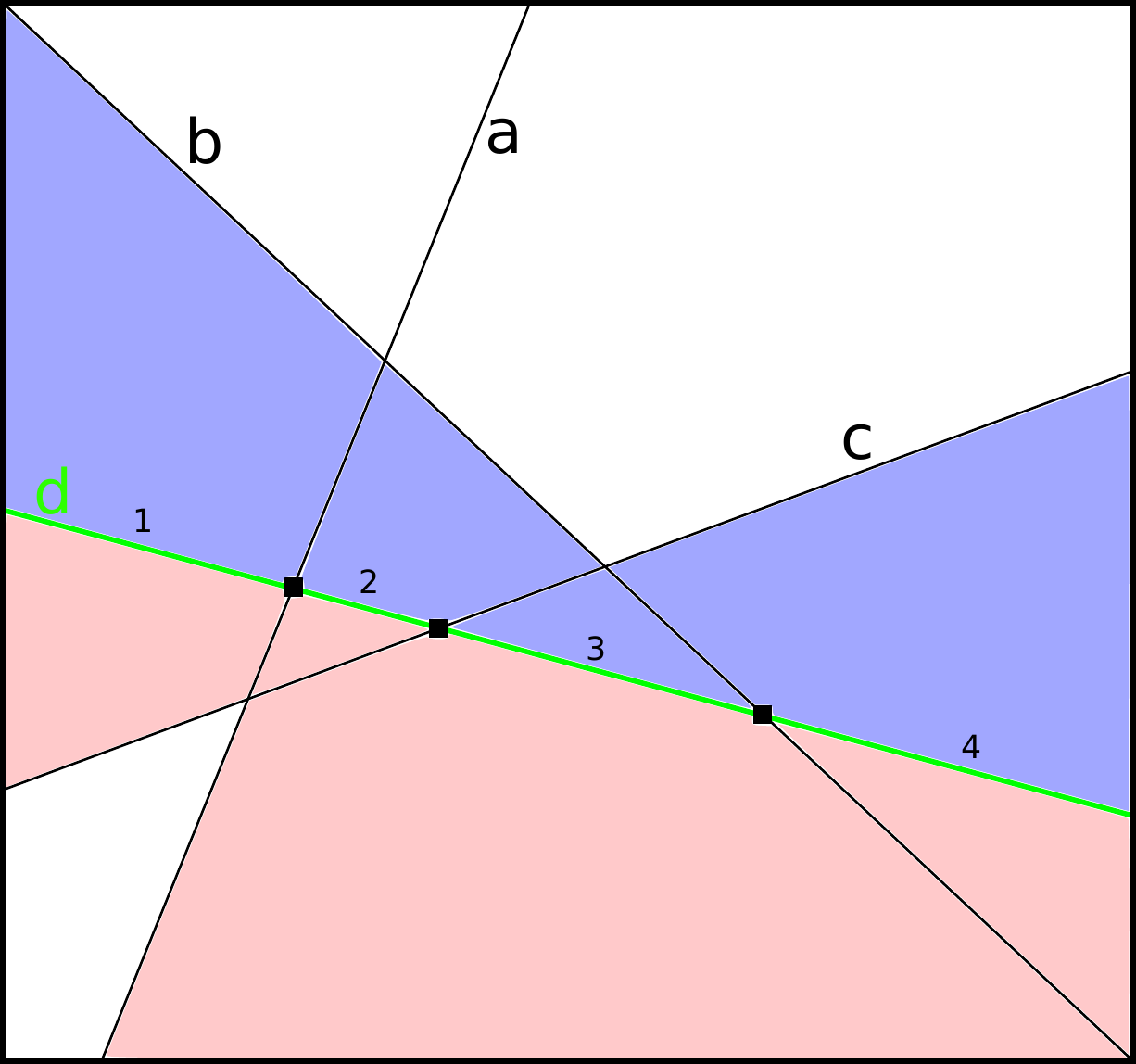}
    \caption{By iteratively introducing lines we can count the maximal number of regions created by $k$ lines. In general positions, the 4th introduced line (d. in greed) will intersect its 3 predecessor in 3 different points. These will create 4 sections, each splitting a region into two (red-blue) hence adding 4 regions to the total count.  }
    \label{fig:nreg}
\end{figure}

\paragraph[Number of regions in a line-arrangement of $n$ lines]{Number of regions in a line-arrangement of $n$ lines \footnote{A \emph{line-arrangement of $n$ lines} is simply a collection of $n$ lines on a plane, which partitions the plane.}}
We start by showing that $r(n)$, the maximal number of regions in $\mathbb{R}^2$ created by a line arrangement of $n$ lines, is $r(n)\leq n^2$. This is based on classic result from computational geometry \citep{zaslavsky1975facing} which we include for completeness. Initially, the entire space is a region. A single line divides the space in two, adding one additional region. What happens as we add additional lines?
The second line intersects\footnote{We assume the added lines are not parallel to any previous line, and do not cross an intersection of previous lines. It is easy to be convinced that such cases will split the space into fewer regions.} with the first, and splits each of the previous regions in two, adding 2 more regions. The third line intersects with both lines, dividing the line into three sections. Each section splits a region, adding 3 more regions. Continuing this way, the $i$th line intersects $i-1$ lines, resulting in $i$ sections, each intersecting a region and thus adding a region. Figure \ref{fig:nreg} illustrates this for the 4th line. We get:
\[
r(n) = 1 + 1 + 2 + 3 + 4 + \ldots + n = 1 + \sum_{i=1}^ni = 1 + \frac{n(n+1)}{2} \leq n^2 \; (\text{for } n>2)
\]

\paragraph{A 1 hidden-layer ReLU network is a line arrangement} 
Consider a network of the form $y = \mathbf{v}(\mathbf{A}x + \mathbf{b})$ where the matrix $\mathbf{A}$ projects the input $x$ to $w$ dimensions, and the vector $\mathbf{v}$ combines them into a weighted sum. The entire input space is linear under this network: the output is linear in the input.\footnote{We can then set a linear classifier by setting a threshold on $y$, this will divide the input space in 2, with a single line.} 
When setting an ReLU activation function after the first layer:
$y = \mathbf{v}\sigma(\mathbf{A}x + \mathbf{b})$ we get a 1-hidden layer ReLU network. For a network with a width $w$ hidden layer ($\mathbf{A}\in\mathbb{R}^{w\times2}$), we get $w$ linear equations, $\mathbf{A}^{(i)}x+\mathbf{b}^{(i)}$ corresponding to $w$ piecewise linear functions: each function has a section where it behaves according to its corresponding equation (the ``active'' section), and a section where it is 0 (the ``rectified'' section).  The input transitions between the active and the rectified sections of function $i$ at the boundary given by $\mathbf{A}^{(i)}x+\mathbf{b}^{(i)}=0$.
Thus, each ReLU neuron corresponds to a line that splits the input space into two: one input region where the neuron is active, and one where it is rectified. Within each region, the behavior of the neuron is linear. For a width $w$ network, we have $w$ such lines --- a line arrangement of $w$ lines. The arrangement splits the space into at most $r(w) < w^2$ convex cells, where each cell corresponds to a set of active neurons. Within each cell, the behavior of the input is linear. Such a cell is called a \emph{linear region}.

\paragraph{Additional Layers} \citep{raghu2017expressive,pascanu2013number} Additional layers further split the linear regions. Consider the network after $d-1$ layers, and a given linear region $R$. Within $R$, the set of active neurons in layers $<d-1$ is constant, and so within the region the next layer computes a linear function of the input. As above, the ReLU activation then again gives $w$ line equations, but this time these equations are only valid within $R$. The next layer than splits $R$ into at most $r(w) \leq w^2$ regions.

\paragraph{Max number of regions in deep networks} \cite{raghu2017expressive}
Consider a network with two hidden layers of width $ w $. The first layer introduced at most $r(w) \leq w^2$ convex regions. As we saw above, for the second layer each region can be split again into at most $r(w)$ regions, resulting in at most $w^2\cdot w^2 = (w^2)^2$ regions.
Applying this recursively, we get that the maximal number of regions in a depth $d$ width $w$ ReLU MLP network is the required bound: $r(w,d) = w^{2d}$. Re-writing $w$ as $2^{log_2w}$ we get: $r(w,d)=2^{log_2w\cdot2d}$.

\section{Conclusion}
We present a depth separation proof for ReLU MLP which is fully self contained and uses only basic mathematical concepts and proof techniques. We believe this work has educational value and new-comers could benefit from its simplicity.
\bibliography{iclr2020_conference}
\bibliographystyle{iclr2020_conference}


\end{document}